\documentclass{article}

\usepackage[utf8]{inputenc}
\usepackage[T1]{fontenc}
\usepackage{hyperref}
\usepackage{url}
\usepackage{booktabs}
\usepackage{amsfonts}
\usepackage{amsmath}
\usepackage{nicefrac}
\usepackage{microtype}
\usepackage{xcolor}
\usepackage{graphicx}
\usepackage{multirow}
\usepackage{subcaption}
\usepackage{enumitem}
\usepackage{wrapfig}
\usepackage{natbib}

\usepackage{float}

\graphicspath{{results/figures/}}

\title{Who Thinks Best Depends on How Long You Let Them:\\ Budget-Dependent Rankings in LLM Evaluation}

\author{Rodrigo Guedes de Souza \and Alison R. Panisson \\
  Federal University of Santa Catarina (UFSC) \\
  Brazil \\
  \texttt{guedes.rodrigo@grad.ufsc.br, alison.panisson@ufsc.br}}

\begin{document}

\maketitle

\begin{abstract}
Standard evaluation of large language models assumes stable model rankings across inference conditions.
We challenge this assumption by varying the \emph{token generation budget}, i.e., the maximum tokens a model may produce, across seven levels (64--4{,}096), evaluating four models on three reasoning benchmarks (56{,}476 inferences).
We report four findings:
\textbf{(i)}~3--19\% of items exhibit non-monotone behavior (accuracy \emph{decreasing} with more budget), even after controlling for truncation, and this phenomenon is model-specific (cross-model overlap: 6--14\%).
\textbf{(ii)}~Model rankings reverse across budgets on all benchmarks ($p {<} 0.01$, McNemar).
\textbf{(iii)}~Oracle analysis reveals model complementarity up to $+27.8$~pp, most pronounced at constrained budgets.
\textbf{(iv)}~A budget-aware router captures 14.1\% of the oracle gap cross-domain; budget features help within-domain ($+1.6$ to $+5.7$~pp) but are domain-specific and hurt transfer ($-1.2$~pp).
These results argue for budget-conditioned evaluation protocols.
\end{abstract}

\section{Introduction}

Leaderboards and benchmark comparisons form the primary evidence base for selecting Large Language Models (LLMs) in practice~\cite{open_llm_leaderboard,chatbot_arena}.
A model declared ``state of the art'' on GSM8K or GPQA is presumed superior across deployment scenarios.
Yet this conclusion rests on an implicit assumption: that model rankings are \emph{invariant} to the inference-time token generation budget, i.e., the maximum number of tokens the model is allowed to produce before its output is truncated or terminated.

This assumption is increasingly untenable.
Recent work on test-time compute scaling~\cite{snell2024scaling,muennighoff2025s1} has shown that allowing models more tokens to ``think'' can substantially improve performance, but the effect varies across models and tasks.
Meanwhile, the ``overthinking'' literature~\cite{chen2024overthinking,sui2025stop} documents cases where additional reasoning tokens \emph{harm} accuracy.
These observations hint at a deeper phenomenon: the evaluation landscape may be fundamentally \emph{budget-dependent}, with model rankings, item difficulty, and model complementarity all varying as a function of the token budget.

We provide the first systematic investigation of this phenomenon.
We evaluate four models spanning 8B to 70B parameters across three reasoning benchmarks at seven token budgets from 64 to 4{,}096 tokens, a total of 56{,}476 individual inferences at temperature $T{=}0$ for full determinism.
Our analysis yields four contributions: 

\begin{enumerate}[leftmargin=*, itemsep=0pt, topsep=0pt]
    \item \textbf{Item-level behavioral taxonomy.} We classify each model--item pair into four behavioral categories under budget variation (always-correct, monotone-increasing, non-monotone, always-wrong). We find that non-monotone behavior, representing performance degradation with increased budget, is not rare (up to 25.8\% of items) and, crucially, is \emph{model-specific}: the same item rarely triggers overthinking across different models (cross-model overlap as low as 9\%).

    \item \textbf{Statistically significant ranking reversals.} The best-performing model changes across budget levels on all three benchmarks. On GSM8K, LLaMA-3.3~70B leads at $b{=}256$ (62.4\%) while GPT-OSS~20B dominates at $b{=}4096$ (94.8\%, $p<0.001$); on GPQA, LLaMA-3~8B ranks first at $b{=}512$ (21.2\%) before GPT-OSS~20B leads nominally at $b{=}4096$ (51.0\% vs.\ 50.0\%, not significant; $n{=}198$). Multiple intermediate reversals are significant (McNemar's $\chi^2$, $p < 0.01$).

    \item \textbf{Oracle gap dynamics.} A per-item oracle ensemble reveals that model complementarity is most valuable under constrained budgets. On GPQA, the oracle exceeds the single best model by 27.8~pp at $b{=}4096$ and exhibits even larger relative gains at lower budgets. On GSM8K, the oracle gap is \emph{non-monotonic}, peaking at $b{=}256$ ($+16.9$~pp) before declining as models converge (Jaccard similarity: $0.048 \to 0.741$).

    \item \textbf{Budget-aware routing proof-of-concept.} We train per-model XGBoost classifiers on text features and budget to predict item-level correctness, then route each item to the model with the highest predicted probability. In a cross-domain evaluation (train on GSM8K$+$MATH-500, test on GPQA), this approach achieves $+2.67$~pp over the best-per-budget baseline (95\% CI~$[0.94, 4.40]$), capturing 14.1\% of the oracle gap. A within-domain ablation reveals that budget features provide $+1.6$ to $+5.7$~pp, yet these patterns are domain-specific and hurt cross-domain transfer by $-1.2$~pp.
\end{enumerate}

\begin{figure}[t]
    \centering
    \includegraphics[width=\textwidth]{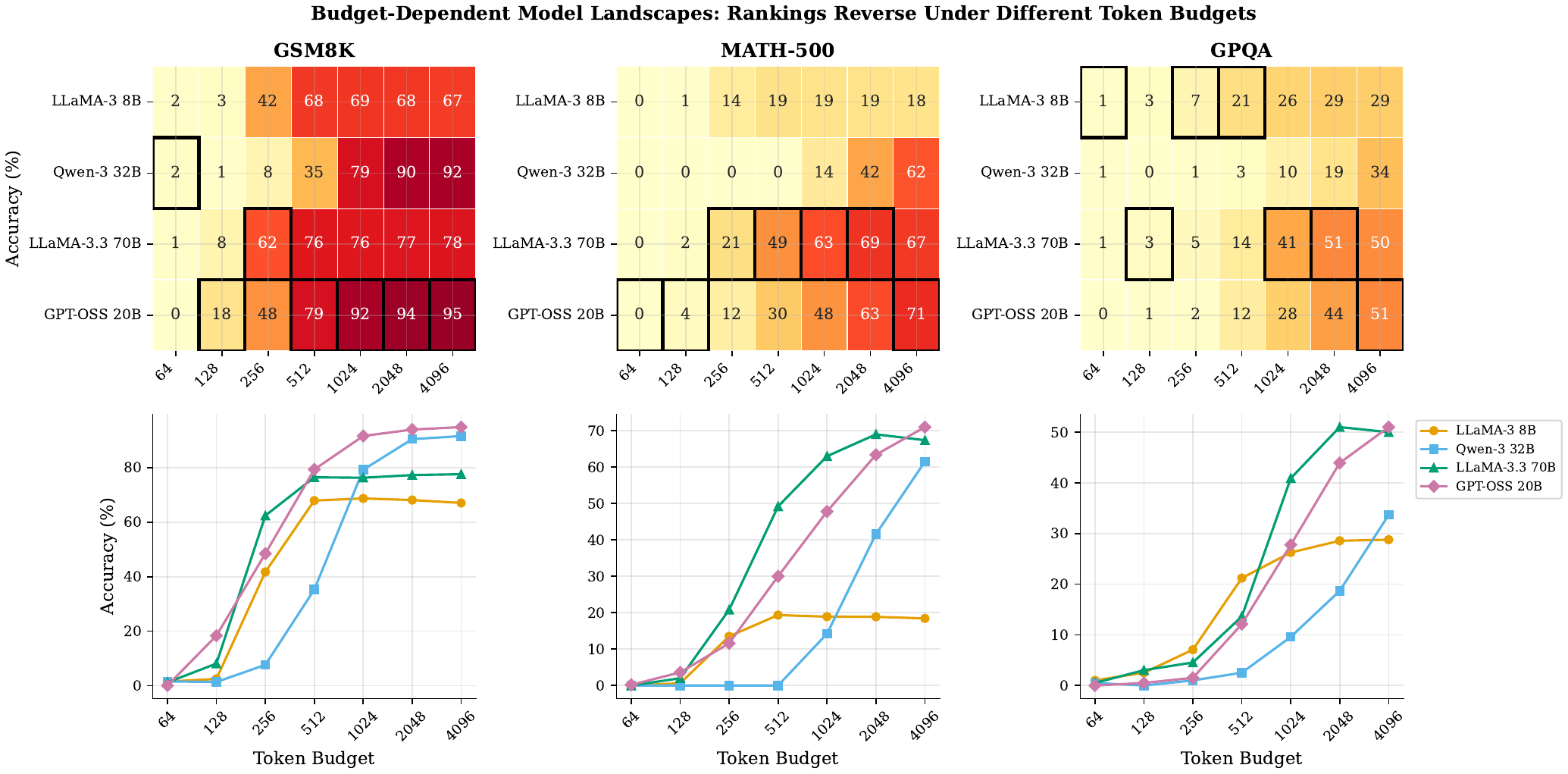}
    \caption{\textbf{Model rankings depend on token budget.} \emph{Top:} Accuracy heatmap across four models and seven budgets on three benchmarks; black borders indicate the best model at each budget. \emph{Bottom:} Corresponding scaling curves showing crossover points where the identity of the top-performing model changes. Rankings reverse on all three benchmarks.}
    \label{fig:teaser}
\end{figure}

Figure~\ref{fig:teaser} illustrates our core finding: model rankings are not stable across token budgets.
These results have direct implications for evaluation practice, model deployment, and the design of routing systems that must operate under heterogeneous compute constraints.

\section{Related Work}

\paragraph{Test-time compute scaling.}
A growing body of work studies how allocating additional computation at inference time affects model performance.
Chain-of-thought prompting~\cite{wei2022chain} implicitly increases the token budget by eliciting step-by-step reasoning.
More recent approaches explicitly scale test-time compute through search~\cite{snell2024scaling}, extended thinking~\cite{muennighoff2025s1}, and budget-forcing mechanisms~\cite{s1_budget_forcing}.
These works generally study a \emph{single} model under varying compute; we study \emph{multiple} models simultaneously, revealing that scaling curves cross and rankings reverse. \vspace{-8pt}

\paragraph{Overthinking and reasoning efficiency.}
Several studies document cases where allowing LLMs more reasoning steps degrades performance. In 
\cite{chen2024overthinking}, the authors identify overthinking as a failure mode of o1-like models, and \cite{sui2025stop} propose early termination strategies.
Our work quantifies this at the item level across multiple models, revealing that overthinking is predominantly a \emph{model-specific} phenomenon rather than an item-inherent property. \vspace{-8pt}

\paragraph{Model routing and selection.}
LLM routing systems~\cite{jiang2023llm_blender,lu2023routing,shnitzer2023routing} aim to select the best model per query based on input features.
Most approaches treat routing as budget-agnostic, optimizing for a single inference condition.
Our budget-aware routing formulation is, to our knowledge, the first to incorporate the token budget as an explicit routing signal, and our SHAP analysis reveals it dominates text-based features. \vspace{-8pt}

\paragraph{Evaluation methodology.}
The reliability of LLM benchmarks has been questioned on multiple fronts: contamination~\cite{sainz2023contamination}, saturation~\cite{kiela2021saturation}, and sensitivity to prompt formatting~\cite{liang2023holistic}.
Our work identifies a new axis of fragility: sensitivity to the token generation budget.
We show that a benchmark's model ranking is not a single stable ordering but a \emph{family} of orderings parameterized by budget.

\section{Experimental Framework}
\label{sec:setup}

\paragraph{Models.}
We evaluate four open-weight reasoning models spanning an order of magnitude in parameter count: \textbf{LLaMA-3~8B}~\cite{llama3}, \textbf{Qwen-3~32B}~\cite{qwen3}, \textbf{LLaMA-3.3~70B}~\cite{llama3}, and \textbf{GPT-OSS~20B} (model ID: \texttt{openai/gpt-oss-20b}, served via Together.ai; see Appendix~\ref{app:details} for full model details).
All models are evaluated with greedy decoding ($T{=}0$) to ensure deterministic outputs. \vspace{-8pt}

\paragraph{Benchmarks.}
We use three reasoning benchmarks of increasing difficulty:
\textbf{GSM8K}~\cite{cobbe2021gsm8k} (1{,}319 grade-school math problems),
\textbf{MATH-500}~\cite{hendrycks2021math} (500 competition-level math problems), and
\textbf{GPQA-Diamond}~\cite{rein2024gpqa} (198 graduate-level science questions).
These span a wide difficulty range: IRT-inspired analysis yields mean easiness scores of 0.507, 0.253, and 0.172, respectively. \vspace{-8pt}

\paragraph{Token budgets.}
We evaluate each model--item pair at seven token budgets $b \in \{64, 128, 256, 512, 1024, 2048, 4096\}$, where $b$ is the \texttt{max\_tokens} parameter.
This yields $4 \times 3 \times 7 = 84$ model--dataset--budget configurations and $4 \times 2{,}017 \times 7 = 56{,}476$ individual inferences. \vspace{-8pt}

\paragraph{Evaluation protocol.}
We extract final answers using regex-based parsing and evaluate via exact match against ground truth.
Correctness is binary: a model scores 1 on item~$i$ at budget~$b$ if and only if it produces the correct final answer within $b$ tokens. \vspace{-8pt}

\paragraph{Truncation and three-tier analysis.}
When a model's generation is cut off at $b$ tokens before producing a final answer, we mark it as truncated (\texttt{finish\_reason = ``length''}) and score it as incorrect.
Truncation rates vary dramatically across models: Qwen-3~32B, which uses internal \texttt{<think>} tokens, remains truncated on 59.7\% of GPQA items even at $b{=}4{,}096$, while LLaMA-3.3~70B drops to 0.5\% at the same budget (see Appendix, Figure~\ref{fig:truncation}).
To disentangle genuine reasoning effects from truncation artifacts, we report a \textbf{three-tier analysis} throughout:
\textbf{(a)}~\emph{all items} (standard scoring),
\textbf{(b)}~\emph{stop-only} (restricting per model to items where that model completed its generation), and
\textbf{(c)}~\emph{common non-truncated} (items where \emph{all four} models completed, enabling paired comparisons on identical item sets).
We note that stop-only accuracy is upward-biased (completed items tend to be easier for that model), while common non-truncated sets can be small at low budgets; we mark results as unreliable when $N < 30$. \vspace{-8pt}

\section{Budget-Dependent Model Behavior}

\subsection{Item-Level Behavioral Taxonomy}
\label{sec:taxonomy}

For each model~$m$ and item~$i$, we observe a binary correctness trajectory across seven budgets: $\mathbf{c}_{m,i} = (c_{m,i}^{b_1}, \ldots, c_{m,i}^{b_7}) \in \{0,1\}^7$.
We classify each trajectory into one of four behavioral categories: 

\begin{itemize}[leftmargin=*, itemsep=1pt, topsep=0pt]
    \item \textbf{Always-correct}: $c_{m,i}^{b} = 1$ for all $b$.  The model solves this item regardless of budget.
    \item \textbf{Monotone-increasing}: the sequence transitions from 0 to 1 and never reverts. More budget always helps.
    \item \textbf{Non-monotone}: the sequence contains at least one $1 \to 0$ transition at increasing budget. The model \emph{loses} a previously correct answer when given more tokens---an ``overthinking'' failure.
    \item \textbf{Always-wrong}: $c_{m,i}^{b} = 0$ for all $b$.  The model fails regardless of budget.
\end{itemize}

\begin{figure}[t]
    \centering
    \includegraphics[width=\textwidth]{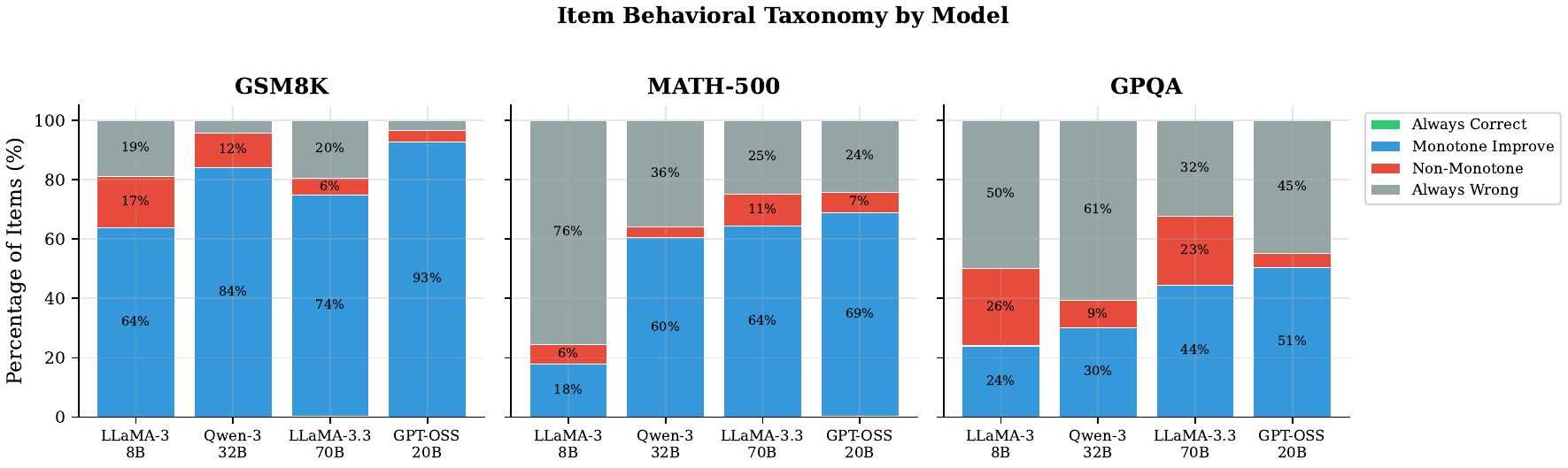}
    \caption{\textbf{Behavioral taxonomy across models and benchmarks.} Each bar decomposes items into four categories based on how correctness evolves with budget. Non-monotone items (red), indicating ``overthinking,'' are non-trivial across all settings and most pronounced on GPQA.}
    \label{fig:taxonomy}
    \vspace{-8pt}
\end{figure}

Figure~\ref{fig:taxonomy} shows the distribution of these categories across all model--dataset combinations.
The monotone-increasing category dominates across all settings, confirming that more budget generally helps.
However, non-monotone items are far from negligible: they represent 3.6\% (GPT-OSS~20B on GSM8K) to 25.8\% (LLaMA-3~8B on GPQA) of items.
GPQA exhibits the highest non-monotone rates across all models, up to 25.8\% for LLaMA-3~8B and 23.2\% for LLaMA-3.3~70B, consistent with the intuition that harder problems are more susceptible to overthinking. \vspace{-8pt}

\begin{figure}[t]
    \centering
    \includegraphics[width=\textwidth]{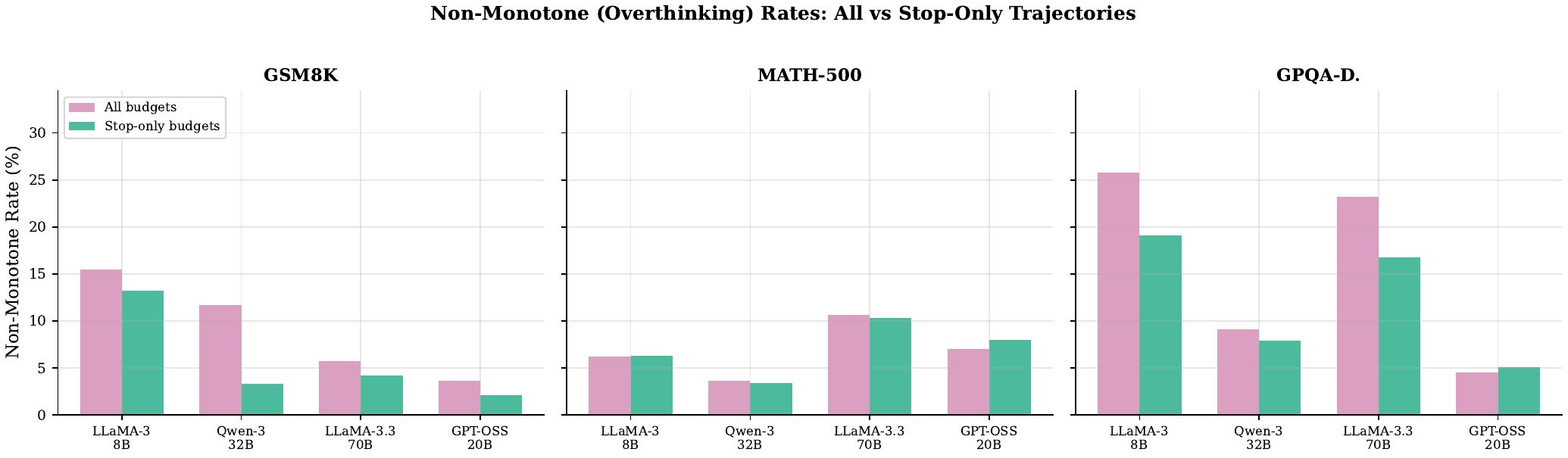}
    \caption{\textbf{Non-monotone (overthinking) rates: all budgets vs.\ stop-only trajectories.} Rates decrease after excluding truncated budgets but remain substantial, particularly on GPQA, confirming that overthinking is a genuine reasoning failure. \vspace{-8pt}}
    \label{fig:taxonomy_filtered}
\end{figure}

\paragraph{Controlling for truncation in the taxonomy.}
Since truncation at lower budgets mechanically creates $1 \to 0$ transitions when a model answers correctly at some budget but is truncated at a later one, we recompute the taxonomy using only the non-truncated portion of each trajectory (budgets where \texttt{finish\_reason = ``stop''}).
Non-monotone rates decrease but remain substantial (Figure~\ref{fig:taxonomy_filtered}): LLaMA-3~8B on GPQA drops from 25.8\% to 19.1\%, LLaMA-3.3~70B on MATH-500 barely changes (10.6\% to 10.3\%), and even Qwen-3~32B on GSM8K retains a 3.3\% non-monotone rate after filtering (down from 11.7\%, where most were truncation artifacts from its verbose \texttt{<think>} tokens).
These results confirm that overthinking is a genuine phenomenon, not a truncation artifact. \vspace{-8pt}

\paragraph{Overthinking is model-specific, not item-inherent.}
A natural question is whether certain items inherently induce overthinking, or whether this is a model-specific phenomenon.
We compute the cross-model overlap: the fraction of non-monotone items that are flagged as non-monotone by at least two of the four models.
On stop-only trajectories, the overlap rates are remarkably low: 10.1\% on GSM8K, 6.4\% on MATH-500, and 13.8\% on GPQA (compared to 16.2\%, 9.7\%, and 27.7\% on all-item trajectories).
Filtering strengthens the claim: 86--94\% of items exhibiting genuine overthinking do so for \emph{only one} model.
The implication is important: overthinking cannot be mitigated by removing ``problematic'' items from benchmarks, because the set of problematic items is model-dependent. \vspace{-8pt}

\paragraph{Where do non-monotone transitions occur?}
Analysis of the budget level at which non-monotone drops happen reveals they are concentrated at high budgets ($b \geq 1024$), suggesting that the additional tokens generated under generous budgets can derail an initially correct reasoning chain (see Appendix, Figure~\ref{fig:transitions}).

\subsection{Ranking Reversals}
\label{sec:reversals}

We now examine whether budget variation leads to changes in which model ranks first, not just changes in absolute performance.
Table~\ref{tab:rankings} reports the best model at selected budget levels along with McNemar test statistics for the significance of the difference with the second-ranked model.

\begin{table}[t]
\centering
\caption{\textbf{Ranking reversals across budgets.} Best model and McNemar's test comparing it to the second-ranked model. The identity of the best model shifts with budget on all three benchmarks. Significance: $^{*}p<0.05$, $^{**}p<0.01$, $^{***}p<0.001$.}
\label{tab:rankings}
\small
\begin{tabular}{@{}llrrcrrc@{}}
\toprule
Dataset & Budget & Best Model & Acc & 2nd Model & Acc & $\chi^2$ & Sig \\
\midrule
\multirow{4}{*}{GSM8K}
& 256  & LLaMA-3.3 70B & 62.4\% & GPT-OSS 20B & 48.6\% & 59.57 & $^{***}$ \\
& 512  & GPT-OSS 20B & 79.2\% & LLaMA-3.3 70B & 76.6\% & 2.78 &  \\
& 1024 & GPT-OSS 20B & 92.3\% & Qwen-3 32B & 80.2\% & 86.12 & $^{***}$ \\
& 4096 & GPT-OSS 20B & 94.9\% & Qwen-3 32B & 91.7\% & 15.86 & $^{***}$ \\
\midrule
\multirow{4}{*}{MATH-500}
& 256  & LLaMA-3.3 70B & 21.3\% & LLaMA-3 8B & 13.5\% & 16.41 & $^{***}$ \\
& 512  & LLaMA-3.3 70B & 49.4\% & GPT-OSS 20B & 30.0\% & 70.51 & $^{***}$ \\
& 1024 & LLaMA-3.3 70B & 63.3\% & GPT-OSS 20B & 48.2\% & 43.81 & $^{***}$ \\
& 4096 & GPT-OSS 20B & 70.8\% & LLaMA-3.3 70B & 67.1\% & 2.54 &  \\
\midrule
\multirow{4}{*}{GPQA}
& 256  & LLaMA-3 8B & 7.1\% & LLaMA-3.3 70B & 4.5\% & 1.07 &  \\
& 512  & LLaMA-3 8B & 21.2\% & LLaMA-3.3 70B & 13.6\% & 4.56 & $^{*}$ \\
& 1024 & LLaMA-3.3 70B & 40.9\% & GPT-OSS 20B & 27.8\% & 9.47 & $^{**}$ \\
& 4096 & GPT-OSS 20B & 51.0\% & LLaMA-3.3 70B & 50.0\% & 0.02 &  \\
\bottomrule
\end{tabular}
\end{table}

Several patterns emerge.
On GSM8K, LLaMA-3.3~70B leads at $b{=}256$ ($\chi^2=59.57$, $p<0.001$), but GPT-OSS~20B overtakes it by $b{=}512$ and maintains the lead through $b{=}4096$ with increasing significance ($\chi^2=15.86$ at $b{=}4096$).
On MATH-500, LLaMA-3.3~70B dominates across mid-range budgets ($b{=}256$ through $b{=}1024$, all $p<0.001$), but GPT-OSS~20B closes the gap and leads at $b{=}4096$---though this final reversal is not significant ($p=0.11$).
On GPQA, the most dramatic trajectory unfolds: the smallest model (LLaMA-3~8B) ranks first at $b{=}256$--$512$, is overtaken by LLaMA-3.3~70B at $b{=}1024$ ($p<0.01$), which is in turn matched by GPT-OSS~20B at $b{=}4096$---though this last difference is not statistically significant ($\chi^2{=}0.02$, $p{=}0.89$, $n{=}198$). Results on GPQA-Diamond should be interpreted with caution due to the small sample size. \vspace{-8pt}

\paragraph{Controlling for truncation.}
Since truncation rates differ across models (e.g., Qwen-3~32B is 59.7\% truncated on GPQA at $b{=}4096$ vs.\ 0.5\% for LLaMA-3.3~70B), ranking reversals could reflect differential truncation rather than genuine reasoning differences.
We repeat the analysis on common non-truncated items, i.e., those where all four models completed their generation, and report paired McNemar tests.
At budgets where common non-truncated sets are sufficiently large ($N \geq 30$), the key findings survive:
on GSM8K at $b{=}4096$ ($N{=}1{,}264$): GPT-OSS~20B leads (95.9\%) over Qwen-3~32B (93.2\%, $\chi^2{=}12.66$, $p{<}0.001$);
on GPQA at $b{=}4096$ ($N{=}71$): GPT-OSS~20B leads (85.9\%) over Qwen-3~32B (64.8\%, $\chi^2{=}9.33$, $p{<}0.01$).

\begin{figure}[t]
    \centering
    \includegraphics[width=\textwidth]{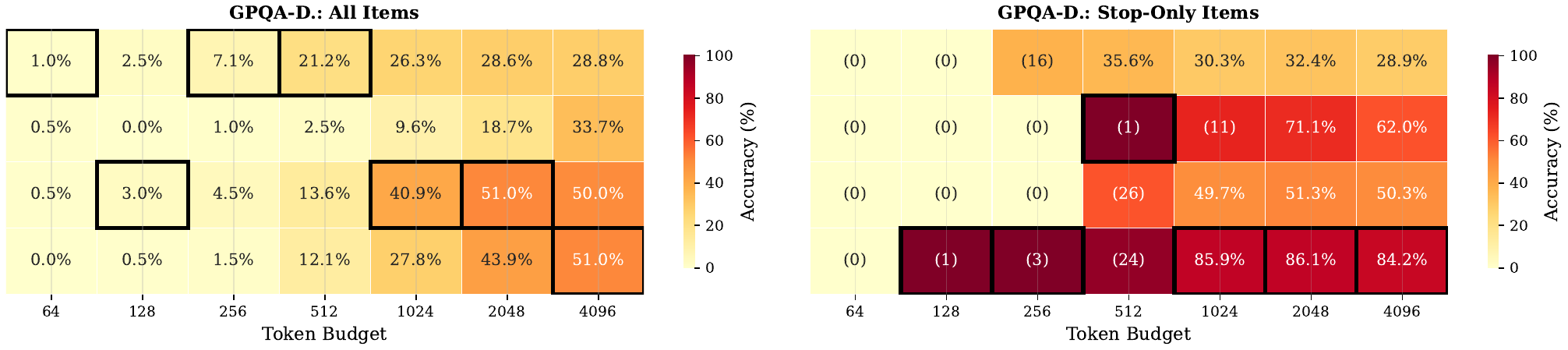}
    \caption{\textbf{All-items vs.\ stop-only accuracy on GPQA.} Left: standard scoring (truncated $=$ incorrect). Right: accuracy restricted to items where each model completed its generation. Parenthesized values indicate small sample sizes ($N < 30$). Black borders mark the best model per budget. Rankings persist at high budgets even under stop-only scoring.}
    \label{fig:dual_heatmap}
\end{figure}

We also observe that \emph{some} all-items rankings are genuine truncation artifacts: on MATH-500 at $b{=}4096$, GPT-OSS~20B leads on all items (71.0\%) but Qwen-3~32B leads on common non-truncated items (86.7\%), reflecting GPT-OSS's higher completion rate rather than superior reasoning.
This nuance strengthens rather than weakens our thesis: the landscape depends on whether one measures ``ability to answer given enough tokens'' or ``ability to answer within the budget,'' and both are legitimate evaluation criteria.
Figure~\ref{fig:dual_heatmap} provides a side-by-side comparison of all-items vs.\ stop-only accuracy.

\subsection{Model Complementarity and Oracle Gap}
\label{sec:oracle}

If different models excel on different items at different budgets, then an oracle that selects the best model per item could substantially outperform any single model.
We quantify this via the \emph{oracle gap}: the difference between a per-item oracle ensemble and the single best model at each budget.

\begin{figure}[t]
    \centering
    \begin{subfigure}[t]{0.49\textwidth}
        \centering
        \includegraphics[width=\textwidth]{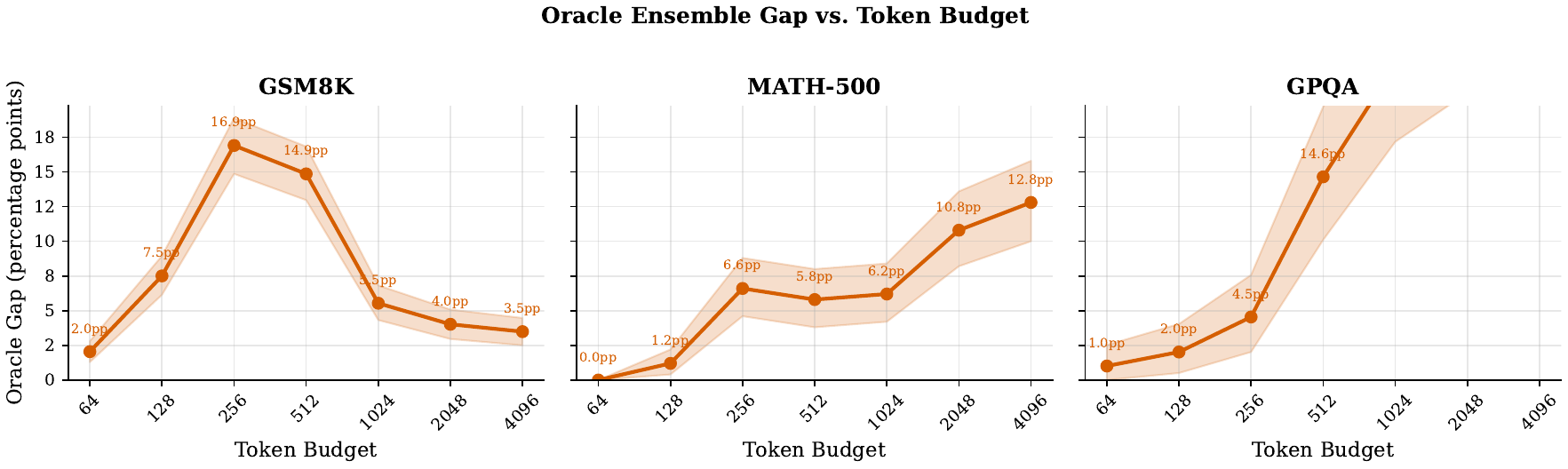}
        \caption{Oracle gap vs.\ token budget.}
        \label{fig:oracle_gap}
    \end{subfigure}
    \hfill
    \begin{subfigure}[t]{0.49\textwidth}
        \centering
        \includegraphics[width=\textwidth]{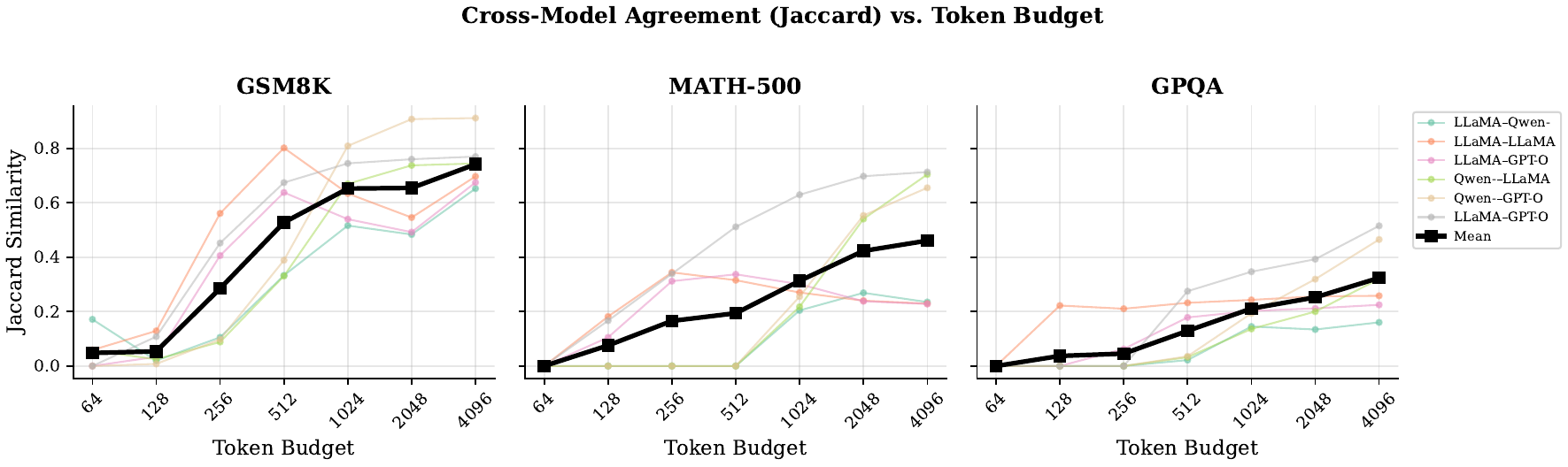}
        \caption{Pairwise Jaccard similarity vs.\ budget.}
        \label{fig:jaccard}
    \end{subfigure}
    \caption{\textbf{Model complementarity varies with budget.} \textbf{(a)}~The oracle gap (oracle ensemble minus best single model) peaks at low-to-moderate budgets for GSM8K but grows monotonically for harder benchmarks. \textbf{(b)}~Mean pairwise Jaccard similarity between models' correct-answer sets starts near zero and increases with budget, but never reaches unity, i.e., models remain complementary even at $b{=}4096$.}
    \label{fig:complementarity}
    \vspace{-8pt}
\end{figure}

Figure~\ref{fig:oracle_gap} plots the oracle gap across budgets for all three benchmarks.
The dynamics differ markedly by dataset difficulty:

\begin{itemize}[leftmargin=*, itemsep=1pt, topsep=0pt]
    \item \textbf{GSM8K} exhibits a \emph{non-monotonic} oracle gap that peaks at $+16.9$~pp around $b{=}256$ before declining to $+3.5$~pp at $b{=}4096$. At moderate budgets, models solve different items, creating maximum complementarity. At high budgets, they converge on the same (easier) items, reducing the benefit of selection.
    \item \textbf{MATH-500} shows a \emph{monotonically increasing} gap, reaching $+12.8$~pp at $b{=}4096$ (CI~$[+10.0, +15.8]$). On this harder benchmark, models remain complementary even at generous budgets.
    \item \textbf{GPQA} displays the largest gap: $+27.8$~pp at $b{=}4096$ (CI~$[+21.7, +33.8]$), indicating that the four models solve highly non-overlapping subsets of these graduate-level questions.
\end{itemize}

\vspace{-8pt}
\paragraph{Convergence without consensus.}
Figure~\ref{fig:jaccard} shows the mean pairwise Jaccard similarity between models' correct-answer sets.
At $b{=}64$, models agree on almost no items ($\bar{J}=0.048$), consistent with the near-random performance at this budget.
As budgets increase, agreement grows substantially: $\bar{J}=0.285$ at $b{=}256$, $0.528$ at $b{=}512$, and $0.741$ at $b{=}4096$.
However, the asymptotic value of 0.741 indicates that even at generous budgets, models solve meaningfully different item subsets, consistent with the persistent oracle gap of 3.5--27.8~pp at $b{=}4096$ across benchmarks. \vspace{-8pt}

\paragraph{Item-level difficulty.}
Across all items and budgets, 1.1\% of GSM8K items, 13.6\% of MATH-500 items, and 10.1\% of GPQA items are answered incorrectly by \emph{all four models at all seven budgets}, representing a hard core of items beyond current model capabilities.
Conversely, only 16.5\% of item--budget pairs see all four models correct, reinforcing that model selection matters.

\section{Budget-Aware Model Routing}
\label{sec:routing}

The oracle gap analysis reveals that an ideal router could achieve 14--28~pp above the best single model.
Can a practical router capture some of this gap?
We design a \emph{budget-aware routing} mechanism that selects the best model for each item, given the token budget as an explicit feature. \vspace{-8pt}

\paragraph{Method.}
For each model $m$, we train an XGBoost binary classifier $f_m(x, b) \to [0,1]$ that predicts the probability that model~$m$ answers item~$x$ correctly at budget~$b$.
Features include: $\log_2(b)$ (budget), surface-level text statistics (character count, word count, number of special characters, presence of LaTeX, word entropy, maximum number magnitude), and 20 PCA-reduced dimensions from sentence embeddings (all-MiniLM-L6-v2).
At inference time, for each item--budget pair, the router selects $m^* = \arg\max_m f_m(x, b)$. \vspace{-8pt}

\paragraph{Baselines.}
We compare against five baselines:
\emph{Random} (uniform selection),
\emph{Largest-Always} (always select GPT-OSS~20B, the largest model),
\emph{Best-Overall} (select the model with highest aggregate accuracy across all budgets),
\emph{Best-Per-Budget} (select the model with highest accuracy at each budget, applied uniformly to all items at that budget), and
\emph{Oracle} (per-item optimal selection with ground truth). \vspace{-8pt}

\paragraph{Cross-domain evaluation.}
To test generalization, we train on GSM8K and MATH-500 and evaluate on GPQA, a domain-transfer setting.
Table~\ref{tab:routing} shows the results.

\begin{table}[t]
\centering
\caption{\textbf{Cross-domain routing results} (train: GSM8K$+$MATH-500, test: GPQA). The per-model scoring router captures 14.1\% of the oracle gap over Best-Per-Budget with a statistically significant improvement. ``Disc.\ subset'' restricts to the 520 out of 1{,}386 item--budget pairs where models disagree.}
\label{tab:routing}
\small
\begin{tabular}{@{}lcccc@{}}
\toprule
Strategy & Accuracy & $\Delta$ vs BPB & Oracle Gap & Disc.\ Subset \\
\midrule
Random & 17.1\% & $-3.2$~pp & --- & 41.0\% \\
Largest-Always & 23.4\% & $+3.1$~pp & --- & 57.7\% \\
Best-Overall & 19.6\% & $-0.7$~pp & --- & 47.5\% \\
Best-Per-Budget & 20.3\% & --- & 0\% & 49.4\% \\
Router-Scoring & 22.9\% & $+2.67$~pp & 14.1\% & 56.5\% \\
\midrule
Oracle & 39.2\% & $+18.9$~pp & 100\% & 100.0\% \\
\bottomrule
\end{tabular}
\end{table}

The Router-Scoring strategy achieves 22.9\% accuracy, a $+2.67$~pp improvement over Best-Per-Budget (95\% bootstrap CI~$[0.94, 4.40]$, significant since the interval excludes zero).
This captures 14.1\% of the oracle gap.
On the discriminative subset (520 items where models disagree), the gain is $+7.12$~pp.

\begin{figure}[t]
    \centering
    \includegraphics[width=0.75\textwidth]{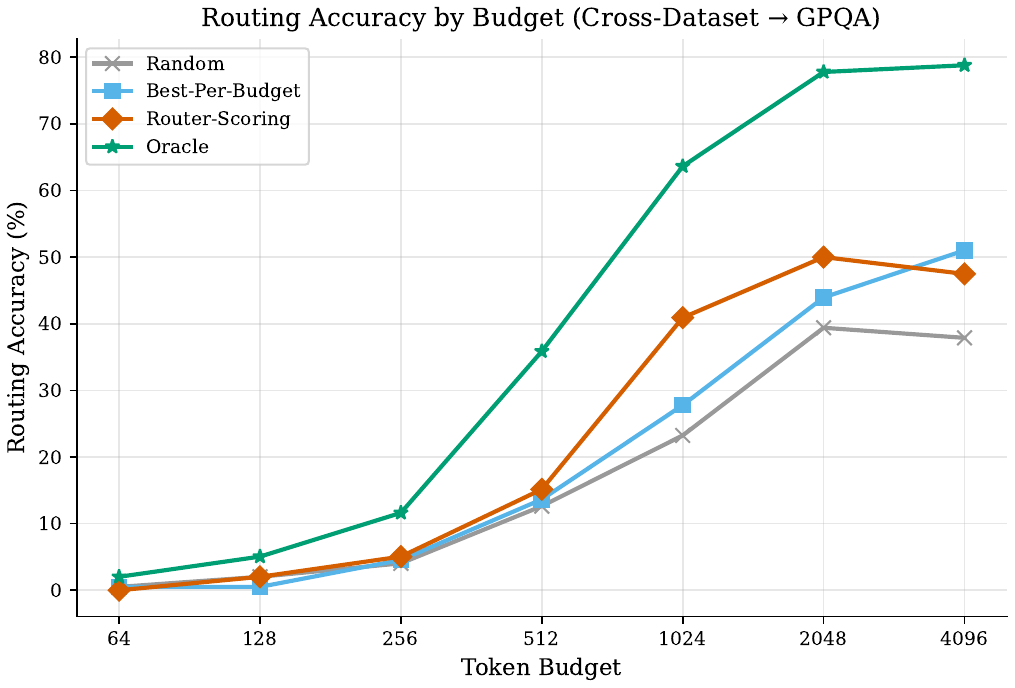}
    \caption{\textbf{Router accuracy by budget level} (cross-domain, GPQA). The router's advantage is concentrated at moderate budgets ($b{=}512$--$2048$), where model complementarity is highest and the oracle gap is largest.}
    \label{fig:routing_budget}
\end{figure}

The per-budget breakdown (Figure~\ref{fig:routing_budget}) reveals that the router's advantage is concentrated at moderate budgets: at $b{=}1024$, it achieves 40.9\% versus 27.8\% for Best-Per-Budget ($+13.1$~pp).
At extreme budgets ($b{=}64$ and $b{=}4096$), the router offers minimal improvement, consistent with the low model differentiation at these extremes.

\vspace{-8pt}
\paragraph{Feature importance.}
SHAP analysis (Figure~\ref{fig:shap}) reveals that the budget feature ($\log_2 b$) dominates all other features by a wide margin: its mean absolute SHAP value (2.21) is $6.1\times$ larger than the next feature (presence of LaTeX: 0.36).
Embedding-derived features contribute modestly (0.10--0.24 each), while text statistics contribute even less.
This confirms that budget is the primary axis of variation in model performance, the core message of this paper. \vspace{-8pt}

\begin{figure}[t]
    \centering
    \includegraphics[width=0.75\textwidth]{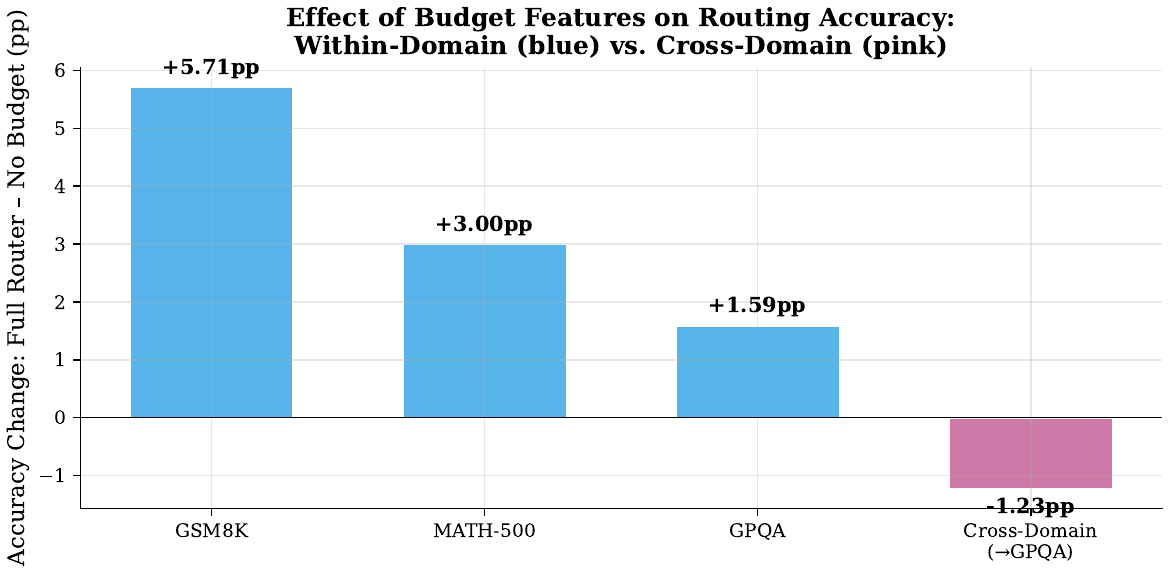}
    \caption{\textbf{Budget feature impact across evaluation settings.} Within-domain, budget is the most valuable feature ($+1.6$ to $+5.7$~pp). Cross-domain, budget features overfit to the training domain and hurt transfer ($-1.2$~pp). This contrast reveals that budget--accuracy patterns are domain-specific.}
    \label{fig:ablation_comparison}
\end{figure}

\paragraph{Ablation study.}
Table~\ref{tab:ablation} reports ablations removing key feature groups in the cross-domain setting.
A notable finding is that removing budget features (``No budget'') yields \emph{higher} cross-domain accuracy (24.2\%) than the full router (22.9\%, $\Delta{=}{-}1.2$~pp).
To determine whether this reflects a global failure of budget features or a domain-transfer artifact, we run a complementary \emph{within-domain} ablation: 5-fold cross-validation on each benchmark separately (Table~\ref{tab:ablation_budget}).
Within-domain, budget features provide a consistent advantage: $+5.7$~pp on GSM8K, $+3.0$~pp on MATH-500, and $+1.6$~pp on GPQA.
The reversal in the cross-domain setting reveals that budget-accuracy mappings are \emph{domain-specific}: the relationship between budget and correctness learned on math problems does not transfer to graduate-level science.
Text features, being domain-agnostic, generalize better across domains---but within a domain, budget remains the dominant routing signal, consistent with the SHAP analysis above.
Figure~\ref{fig:ablation_comparison} visualizes this contrast.

\begin{table}[t]
\centering
\caption{\textbf{Cross-domain ablation} (train: GSM8K$+$MATH-500, test: GPQA). Removing budget features \emph{improves} cross-domain accuracy, suggesting that budget--accuracy patterns are domain-specific.}
\label{tab:ablation}
\small
\begin{tabular}{@{}lc@{}}
\toprule
Configuration & Accuracy \\
\midrule
Random & 17.1\% \\
Best-Per-Budget & 20.3\% \\
Text stats only (no budget, no embeddings) & 21.8\% \\
No budget & 24.2\% \\
No embeddings & 21.3\% \\
Full Router & 22.9\% \\
Oracle & 39.2\% \\
\bottomrule
\end{tabular}
\end{table}

\begin{table}[t]
\centering
\caption{\textbf{Budget feature impact: within-domain vs.\ cross-domain.} Budget features consistently help within-domain (5-fold CV) but hurt cross-domain transfer, confirming that budget--accuracy patterns are domain-specific rather than universally transferable.}
\label{tab:ablation_budget}
\small
\begin{tabular}{@{}lccc@{}}
\toprule
Setting & Full Router & No Budget & $\Delta$ Budget \\
\midrule
GSM8K (within-domain) & 64.9\% & 59.2\% & $+5.7$~pp \\
MATH-500 (within-domain) & 37.5\% & 34.5\% & $+3.0$~pp \\
GPQA (within-domain) & 24.0\% & 22.4\% & $+1.6$~pp \\
\midrule
Cross$\to$GPQA & 22.9\% & 24.2\% & $-1.2$~pp \\
\bottomrule
\end{tabular}
\end{table}

\section{Discussion}
\label{sec:discussion}

\paragraph{Implications for evaluation.}
Our findings challenge the common practice of reporting a single accuracy number per benchmark.
Since model rankings depend on the token budget, a benchmark's ranking is not a fixed ordering but a \emph{family} of orderings parameterized by $b$.
We advocate for budget-conditioned evaluation: reporting accuracy at multiple budget levels and explicitly stating the budget used.
This is especially critical for constrained deployment scenarios (mobile, edge, real-time) where generous budgets are infeasible.\vspace{-8pt}

\paragraph{Implications for routing.}
The large oracle gaps ($+3.5$ to $+27.8$~pp) indicate that substantial gains are available from intelligent model selection.
Our routing experiments reveal a nuanced picture: budget is the dominant within-domain feature ($+1.6$ to $+5.7$~pp), yet budget--accuracy patterns are domain-specific and hurt cross-domain transfer ($-1.2$~pp).
This suggests that practical routing systems should treat budget as a powerful but non-transferable signal, complementing it with domain-agnostic features or domain adaptation techniques.
Future routing systems may additionally benefit from model-internal signals (e.g., logit entropy, hidden-state representations) rather than text features alone.\vspace{-8pt}

\paragraph{The truncation confound.}
Truncation is a significant confound when varying token budgets: at low budgets, most models are truncated on most items, and truncation is near-perfectly correlated with incorrect answers.
We have addressed this through a three-tier analysis (Section~\ref{sec:reversals}): all items, stop-only per model, and common non-truncated items with paired McNemar tests.
The key findings, non-monotone behavior (Section~\ref{sec:taxonomy}), ranking reversals, and oracle gaps, all persist under truncation control, though some specific rankings are revealed to be truncation artifacts (e.g., MATH-500 at $b{=}4096$; see Section~\ref{sec:reversals}).
Notably, non-monotone rates decrease by only 1--8~pp after filtering truncated budgets, confirming that overthinking is predominantly a genuine reasoning failure.
We further validate this with a direct search for items where a model is correct at a low budget but wrong at a higher budget, both with \texttt{finish\_reason=stop}: across all models and datasets, we identify 1{,}193 such pairs (Appendix~\ref{app:overthinking}).
Future work should explore ``budget-forcing'' techniques~\cite{s1_budget_forcing} that encourage models to produce complete answers within the budget, rather than simply truncating.\vspace{-8pt}

\paragraph{Limitations.}
Our study has a few limitations.
\textbf{(i)}~We evaluate only four models; expanding to a broader set would strengthen the generality of our findings.
\textbf{(ii)}~Our token budgets are logarithmically spaced; finer-grained budgets might reveal additional structure.
\textbf{(iii)}~Our routing features are limited to surface-level text properties and static embeddings; richer features could capture more of the oracle gap.
\textbf{(iv)}~We focus on reasoning benchmarks with clear correct answers; generalization to open-ended tasks remains untested.
\textbf{(v)}~The 19~pp gap between our router and the oracle on GPQA suggests that most model complementarity remains unexploited, which is a compelling direction for future work.
\textbf{(vi)}~We deliberately exclude dedicated reasoning models (o1, DeepSeek-R1, QwQ) because their dual-stream architecture, allocating tokens between internal ``thinking'' and visible output, changes the semantics of \texttt{max\_tokens}. Controlling \texttt{max\_tokens} restricts visible output but may not constrain internal deliberation, making budget comparisons non-equivalent. Extending our framework to reasoning models is an important direction for future work.

\section{Conclusion}

We have shown that the evaluation landscape of LLMs is fundamentally budget-dependent.
Model rankings, item difficulty, and model complementarity all vary as a function of the token generation budget.
These findings have three actionable implications: (1)~benchmarks should adopt budget-conditioned evaluation protocols, reporting rankings at multiple budget levels; (2)~model selection and routing systems should incorporate budget as a first-class signal; and (3)~the substantial oracle gaps we document ($+3.5$ to $+27.8$~pp) represent a concrete opportunity for ensemble and routing methods to exploit model complementarity.

The question ``which model is best?'' has no single answer, it depends on how long you let them think.

\bibliographystyle{abbrv}

\begin{thebibliography}{20}

\bibitem{open_llm_leaderboard}
Open LLM Leaderboard.
\newblock \url{https://huggingface.co/spaces/open-llm-leaderboard/open_llm_leaderboard}, 2024.

\bibitem{chatbot_arena}
L.~Zheng, W.-L. Chiang, et~al.
\newblock Judging LLM-as-a-Judge with MT-Bench and Chatbot Arena.
\newblock In \emph{NeurIPS}, 2023.

\bibitem{snell2024scaling}
C.~Snell, J.~Lee, K.~Xu, and A.~Kumar.
\newblock Scaling LLM Test-Time Compute Optimally Can be More Effective than Scaling Model Parameters.
\newblock \emph{arXiv:2408.03314}, 2024.

\bibitem{muennighoff2025s1}
N.~Muennighoff, Z.~Yang, et~al.
\newblock s1: Simple Test-Time Scaling.
\newblock \emph{arXiv:2501.19393}, 2025.

\bibitem{s1_budget_forcing}
S.~Aggarwal, Y.~Arora, and A.~Goyal.
\newblock L1: Controlling How Long A Reasoning Model Thinks With Reinforcement Learning.
\newblock \emph{arXiv:2503.04697}, 2025.

\bibitem{wei2022chain}
J.~Wei, X.~Wang, D.~Schuurmans, et~al.
\newblock Chain-of-Thought Prompting Elicits Reasoning in Large Language Models.
\newblock In \emph{NeurIPS}, 2022.

\bibitem{chen2024overthinking}
X.~Chen, Z.~Xu, et~al.
\newblock Do NOT Think That Much for 2+3=? On the Overthinking of o1-Like LLMs.
\newblock \emph{arXiv:2412.21187}, 2024.

\bibitem{sui2025stop}
Y.~Sui, H.~Yu, et~al.
\newblock Stop Overthinking: A Survey on Efficient Reasoning for Large Language Models.
\newblock \emph{arXiv:2503.16419}, 2025.

\bibitem{jiang2023llm_blender}
D.~Jiang, X.~Ren, and B.~Y. Lin.
\newblock LLM-Blender: Ensembling Large Language Models with Pairwise Ranking and Generative Fusion.
\newblock In \emph{ACL}, 2023.

\bibitem{lu2023routing}
K.~Lu, H.~Yuan, et~al.
\newblock Routing to the Expert: Efficient Reward-guided Ensemble of Large Language Models.
\newblock In \emph{NAACL}, 2024.

\bibitem{shnitzer2023routing}
T.~Shnitzer, A.~Ou, et~al.
\newblock Large Language Model Routing with Benchmark Datasets.
\newblock \emph{arXiv:2309.15789}, 2023.

\bibitem{sainz2023contamination}
O.~Sainz, J.~Campos, et~al.
\newblock NLP Evaluation in Trouble: On the Need to Measure LLM Data Contamination for each Benchmark.
\newblock In \emph{EMNLP Findings}, 2023.

\bibitem{kiela2021saturation}
D.~Kiela, M.~Bartolo, et~al.
\newblock Dynabench: Rethinking Benchmarking in NLP.
\newblock In \emph{NAACL}, 2021.

\bibitem{liang2023holistic}
P.~Liang, R.~Bommasani, et~al.
\newblock Holistic Evaluation of Language Models.
\newblock \emph{Annals of the New York Academy of Sciences}, 2023.

\bibitem{llama3}
Meta AI.
\newblock The LLaMA 3 Herd of Models.
\newblock \emph{arXiv:2407.21783}, 2024.

\bibitem{qwen3}
Qwen Team.
\newblock Qwen3 Technical Report.
\newblock \emph{arXiv:2505.09388}, 2025.

\bibitem{cobbe2021gsm8k}
K.~Cobbe, V.~Kosaraju, et~al.
\newblock Training Verifiers to Solve Math Word Problems.
\newblock \emph{arXiv:2110.14168}, 2021.

\bibitem{hendrycks2021math}
D.~Hendrycks, C.~Burns, et~al.
\newblock Measuring Mathematical Problem Solving With the MATH Dataset.
\newblock In \emph{NeurIPS}, 2021.

\bibitem{rein2024gpqa}
D.~Rein, B.~L. Hou, et~al.
\newblock GPQA: A Graduate-Level Google-Proof Q\&A Benchmark.
\newblock In \emph{ICLR}, 2024.

\end{thebibliography}

\newpage
\appendix
\section{Additional Experimental Details}
\label{app:details}

\paragraph{Model details.}
All models are evaluated via their standard chat/instruct variants through a unified API.
Table~\ref{tab:model_details} provides full identification for all models.
The models span a deliberate range of architectures and parameter counts to test whether budget-dependence is a universal phenomenon or model-specific.
Temperature is set to 0.0 for all models to ensure deterministic outputs and exact reproducibility.

\begin{table}[h]
\centering
\caption{\textbf{Full model identification.}}
\label{tab:model_details}
\small
\begin{tabular}{@{}llll@{}}
\toprule
Paper Name & Model ID & Provider(s) & Params \\
\midrule
LLaMA-3 8B & \texttt{meta-llama/Llama-3.1-8B-Instruct} & Groq, Cerebras, SambaNova & 8B \\
Qwen-3 32B & \texttt{Qwen3-32B} & Groq, SambaNova & 32B \\
LLaMA-3.3 70B & \texttt{meta-llama/Llama-3.3-70B-Instruct-Turbo} & Together.ai & 70B \\
GPT-OSS 20B & \texttt{openai/gpt-oss-20b} & Together.ai & 20B \\
\bottomrule
\end{tabular}
\end{table}

\paragraph{Compute resources.}
Inference was performed using cloud API endpoints. Total inference count: 56{,}476 individual API calls (4~models $\times$ 2{,}017~items $\times$ 7~budgets). Routing experiments used scikit-learn and XGBoost on a single CPU. Sentence embeddings were computed with all-MiniLM-L6-v2 (22M parameters).

\section{Full Accuracy Tables}
\label{app:accuracy}

Table~\ref{tab:full_gsm8k} shows accuracy and 95\% bootstrap confidence intervals for all model--budget combinations on GSM8K.

\begin{table}[h]
\centering
\caption{\textbf{Full accuracy with 95\% bootstrap CIs on GSM8K.}}
\label{tab:full_gsm8k}
\small
\begin{tabular}{@{}lcccccc@{}}
\toprule
Model & $b{=}64$ & $b{=}256$ & $b{=}512$ & $b{=}1024$ & $b{=}2048$ & $b{=}4096$ \\
\midrule
LLaMA-3 8B & \hphantom{0}1.5\tiny{$\pm$0.7} & 41.7\tiny{$\pm$2.6} & 67.9\tiny{$\pm$2.5} & 68.7\tiny{$\pm$2.8} & 68.1 & 67.0\tiny{$\pm$2.6}\\
Qwen-3 32B & \hphantom{0}1.6\tiny{$\pm$0.7} & \hphantom{0}7.7\tiny{$\pm$1.5} & 35.3\tiny{$\pm$2.6} & 79.2\tiny{$\pm$2.2} & 90.4 & 91.5\tiny{$\pm$1.5}\\
LLaMA-3.3 70B & \hphantom{0}1.2\tiny{$\pm$0.6} & 62.4\tiny{$\pm$2.6} & 76.4\tiny{$\pm$2.3} & 76.3\tiny{$\pm$2.3} & 77.6 & 77.6\tiny{$\pm$2.3}\\
GPT-OSS 20B & \hphantom{0}0.1\tiny{$\pm$0.1} & 48.4\tiny{$\pm$2.7} & 79.3\tiny{$\pm$2.2} & 91.6\tiny{$\pm$1.5} & 93.9 & 94.8\tiny{$\pm$1.2}\\
\bottomrule
\end{tabular}
\end{table}

\section{Truncation Analysis}
\label{app:truncation}

\begin{figure}[H]
    \centering
    \includegraphics[width=0.7\textwidth]{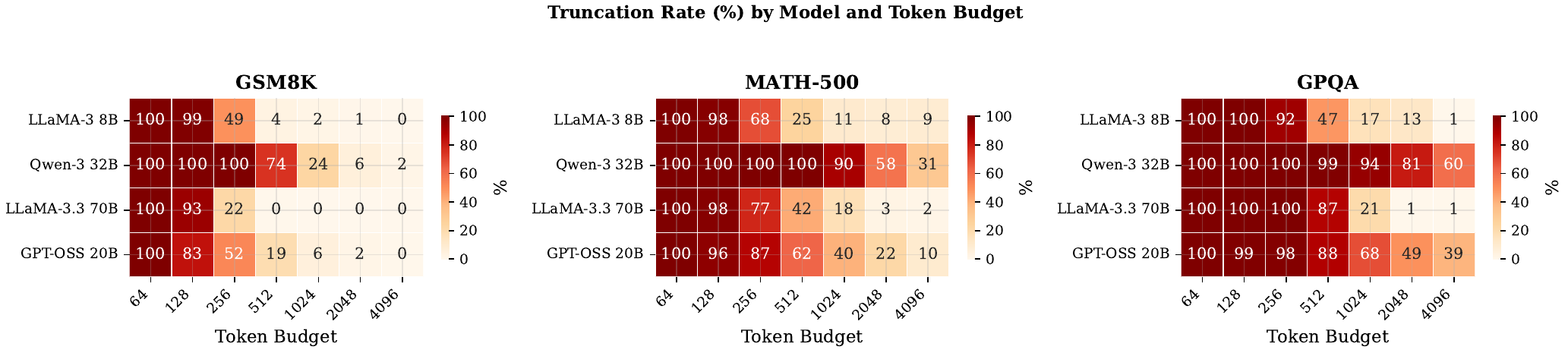}
    \caption{\textbf{Truncation rate by model, budget, and dataset.} Qwen-3~32B experiences the highest truncation rates across all budgets, while LLaMA-3.3~70B shows the fastest decline. At $b{=}4096$, truncation rates are near zero for all models except Qwen-3~32B (15\% on GSM8K, higher on MATH-500 and GPQA).}
    \label{fig:truncation}
\end{figure}

\section{Non-Monotone Transition Analysis}
\label{app:transitions}

\begin{figure}[H]
    \centering
    \includegraphics[width=0.7\textwidth]{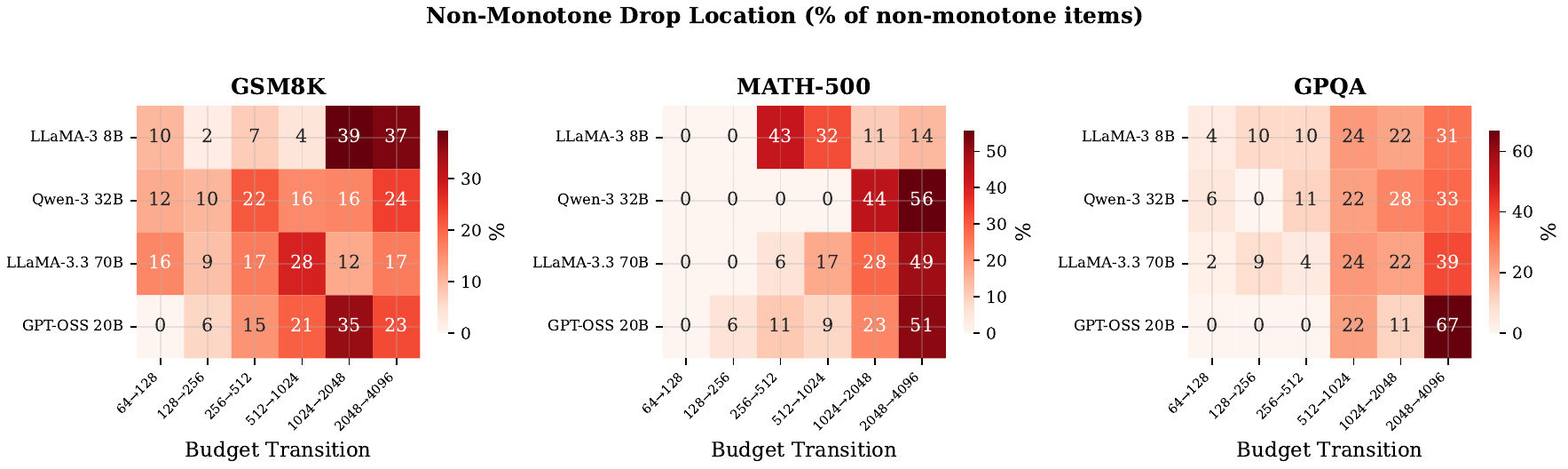}
    \caption{\textbf{Distribution of non-monotone transitions by budget level.} The heatmap shows at which budget level non-monotone items first lose a previously correct answer. Most transitions are concentrated at high budgets ($b \geq 1024$), especially for GSM8K.}
    \label{fig:transitions}
\end{figure}

\section{SHAP Feature Importance}
\label{app:shap}

\begin{figure}[H]
    \centering
    \includegraphics[width=0.7\textwidth]{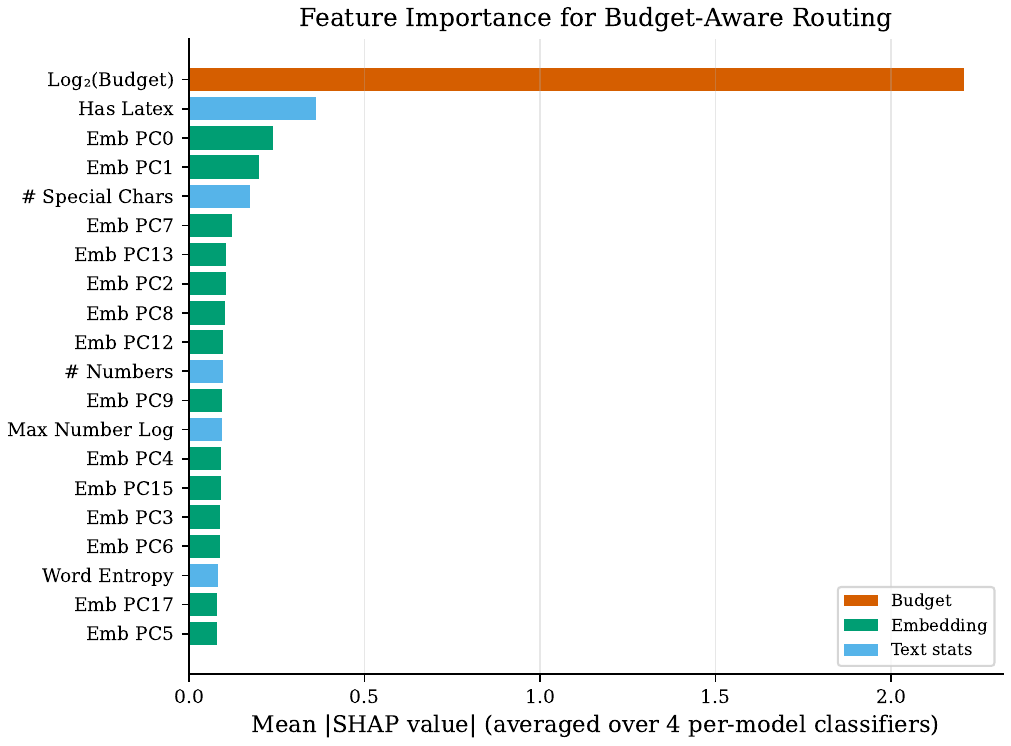}
    \caption{\textbf{SHAP feature importance for the per-model scoring router.} Budget ($\log_2 b$) dominates with a mean absolute SHAP value of 2.21, approximately $6\times$ larger than the next feature (presence of LaTeX: 0.36). Text features and embeddings contribute modestly.}
    \label{fig:shap}
\end{figure}

\section{IRT Difficulty Analysis}
\label{app:irt}

\begin{figure}[H]
    \centering
    \includegraphics[width=0.7\textwidth]{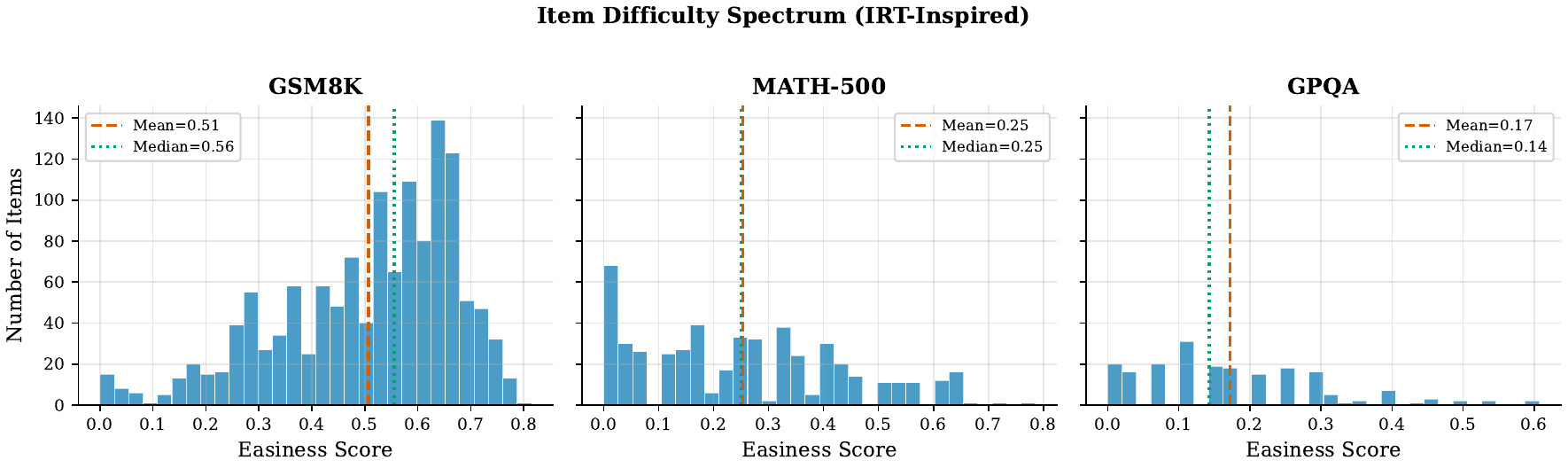}
    \caption{\textbf{IRT-inspired difficulty distribution across benchmarks.} Easiness is computed as the fraction of model--budget combinations that answer each item correctly. GSM8K items cluster around easiness 0.5 (mean: 0.507), MATH-500 around 0.25 (mean: 0.253), and GPQA around 0.17 (mean: 0.172), confirming the intended difficulty gradient.}
    \label{fig:irt}
\end{figure}

\section{Genuine Overthinking Examples}
\label{app:overthinking}

To confirm that non-monotone behavior is not merely a truncation artifact, we identify items where a model answers correctly at a low budget $b_l$ but incorrectly at a higher budget $b_h > b_l$, with \texttt{finish\_reason=stop} at \emph{both} budgets (i.e., neither response was truncated).
Table~\ref{tab:overthinking_counts} reports the number of unique items exhibiting this pattern per model and dataset.
Across the four models and three benchmarks, we find 1{,}193 such (low-budget, high-budget) pairs---concrete evidence that overthinking is a genuine reasoning failure, not an artifact of output truncation.

LLaMA-3~8B exhibits the most overthinking instances (550 on GSM8K alone), consistent with its high non-monotone rate in the behavioral taxonomy (Section~\ref{sec:taxonomy}).
Qwen-3~32B shows the fewest (65 on GSM8K, 4 on GPQA), despite its high truncation rate, suggesting that its non-monotone behavior is more often truncation-driven.

\begin{table}[h]
\centering
\caption{\textbf{Genuine overthinking instances} (correct$\to$wrong, both with \texttt{finish\_reason=stop}). Counts show the number of unique items where a model answers correctly at a lower budget but incorrectly at a higher budget, with no truncation at either level.}
\label{tab:overthinking_counts}
\small
\begin{tabular}{@{}lccc@{}}
\toprule
Model & GSM8K & MATH-500 & GPQA \\
\midrule
LLaMA-3 8B & 550 & 88 & 74 \\
Qwen-3 32B & 65 & 9 & 4 \\
LLaMA-3.3 70B & 159 & 93 & 45 \\
GPT-OSS 20B & 46 & 52 & 8 \\
\bottomrule
\end{tabular}
\end{table}

\section{Prompt Templates}
\label{app:prompts}

We use the following prompt templates, adapted per benchmark. For GSM8K and MATH-500 (numerical answer):

\begin{quote}
\small
\texttt{Solve the following problem step by step. Show your complete reasoning.}\\[4pt]
\texttt{Problem: \{question\}}\\[4pt]
\texttt{After your reasoning, provide your final answer on a new line in the exact format:}\\
\texttt{\#\#\#\# [your answer]}
\end{quote}

For GPQA-Diamond (multiple choice):

\begin{quote}
\small
\texttt{Answer the following question by reasoning step by step.}\\[4pt]
\texttt{Question: \{question\}}\\[4pt]
\texttt{Options:}\\
\texttt{(A) \{option\_a\}}\\
\texttt{(B) \{option\_b\}}\\
\texttt{(C) \{option\_c\}}\\
\texttt{(D) \{option\_d\}}\\[4pt]
\texttt{After your reasoning, provide your final answer on a new line in the exact format:}\\
\texttt{\#\#\#\# [A/B/C/D]}
\end{quote}

\paragraph{Answer parsing.}
For numerical benchmarks (GSM8K, MATH-500), we parse the \texttt{\#\#\#\# [answer]} pattern and compare numerically with tolerance $10^{-6}$.
Format variations (e.g., $\frac{1}{2}$ vs.\ 0.5) are handled via float conversion with LaTeX-aware parsing.
For GPQA, we extract a single letter (A--D) using a cascade of regex patterns applied to the model output.

\end{document}